\pdfoutput=1

\documentclass[11pt]{article}

\usepackage{naacl2021}

\usepackage{times}
\usepackage{latexsym}
\usepackage{graphicx}
\usepackage{float}

\usepackage[T1]{fontenc}

\usepackage[utf8]{inputenc}

\usepackage{microtype}
\usepackage[flushleft]{threeparttable}
\usepackage{booktabs}

\usepackage{xspace}
\newcommand{\pawls}{\texttt{PAWLS}\xspace}

\title{\pawls: PDF Annotation With Labels and Structure}

\author{Mark Neumann 
   \And
  Zejiang Shen \\\\
    Allen Institute for Artificial Intelligence \\
    \texttt{\{markn,shannons,sams\}@allenai.org}
  \And
  Sam Skjonsberg \\
}

\begin{document}
\maketitle
\begin{abstract}

Adobe's Portable Document Format (PDF) is a popular way of distributing view-only documents with a rich visual markup. This presents a challenge to NLP practitioners who wish to use the information contained within PDF documents for training models or data analysis, because annotating these documents is difficult. In this paper, we present PDF Annotation with Labels and Structure (\pawls), a new annotation tool designed specifically for the PDF document format. \pawls is particularly suited for mixed-mode annotation and scenarios in which annotators require extended context to annotate accurately. \pawls supports span-based textual annotation, N-ary relations and freeform, non-textual bounding boxes, all of which can be exported in convenient formats for training multi-modal machine learning models. A read-only \pawls server is available at \url{https://pawls.apps.allenai.org/} \footnote{Please see Appendix \ref{demo_access} for instructions on accessing the demo.} and the source code is available at \url{https://github.com/allenai/pawls}.
\end{abstract}

\section{Introduction}

Authors of Natural Language Processing technology rely on access to gold standard annotated data for training and evaluation of learning algorithms. Despite successful attempts to create machine readable document formats such as XML and HTML, the Portable Document Format (PDF) is still widely used for read only documents which require visual markup, across domains such as scientific publishing, law and government. This presents a challenge to NLP practitioners, as the PDF format does not contain exhaustive markup information, making it difficult to extract semantically meaningful regions from a PDF. Annotating text extracted from PDFs in a plaintext format is difficult, because the extracted text stream lacks any organisation or markup, such as paragraph boundaries, figure placement and page headers/footers.

Existing popular annotation tools such as BRAT \cite{stenetorp-etal-2012-brat} focus on annotation of user provided plain text in a web browser specifically designed for annotation only. For many labeling tasks, this format is exactly what is required. However, as the scope and ability of natural language processing technology goes beyond purely textual processing due in part to recent advances in large language models \cite[][\textit{inter alia}]{peters-etal-2018-deep, devlin-etal-2019-bert}, the context and media in which datasets are created must evolve as well.

In addition, the quality of both data collection and evaluation methodology is highly dependent on the particular annotation/evaluation context in which the data being annotated is viewed \cite{joseph-etal-2017-constance, laubli-etal-2018-machine}. Annotating data directly on top of a PDF canvas allows naturally occurring text to be collected in addition to it to being by annotators in it's original context - that of the PDF itself. 

To address the need for an annotation tool that goes beyond plaintext data, we present a new annotation tool called \pawls (PDF Annotation With Labels and Structure). In this paper, we discuss some of the PDF specific design choices in \pawls, including automatic bounding box uniformity, free-form annotations for non-textual image regions and scale/dimension agnostic bounding box storage. We report agreement statistics from an initial round of labelling during the creation of a PDF structure parsing dataset for which \pawls was originally designed.

\begin{figure*}[ht]
    \centering
    \includegraphics[width=\textwidth]{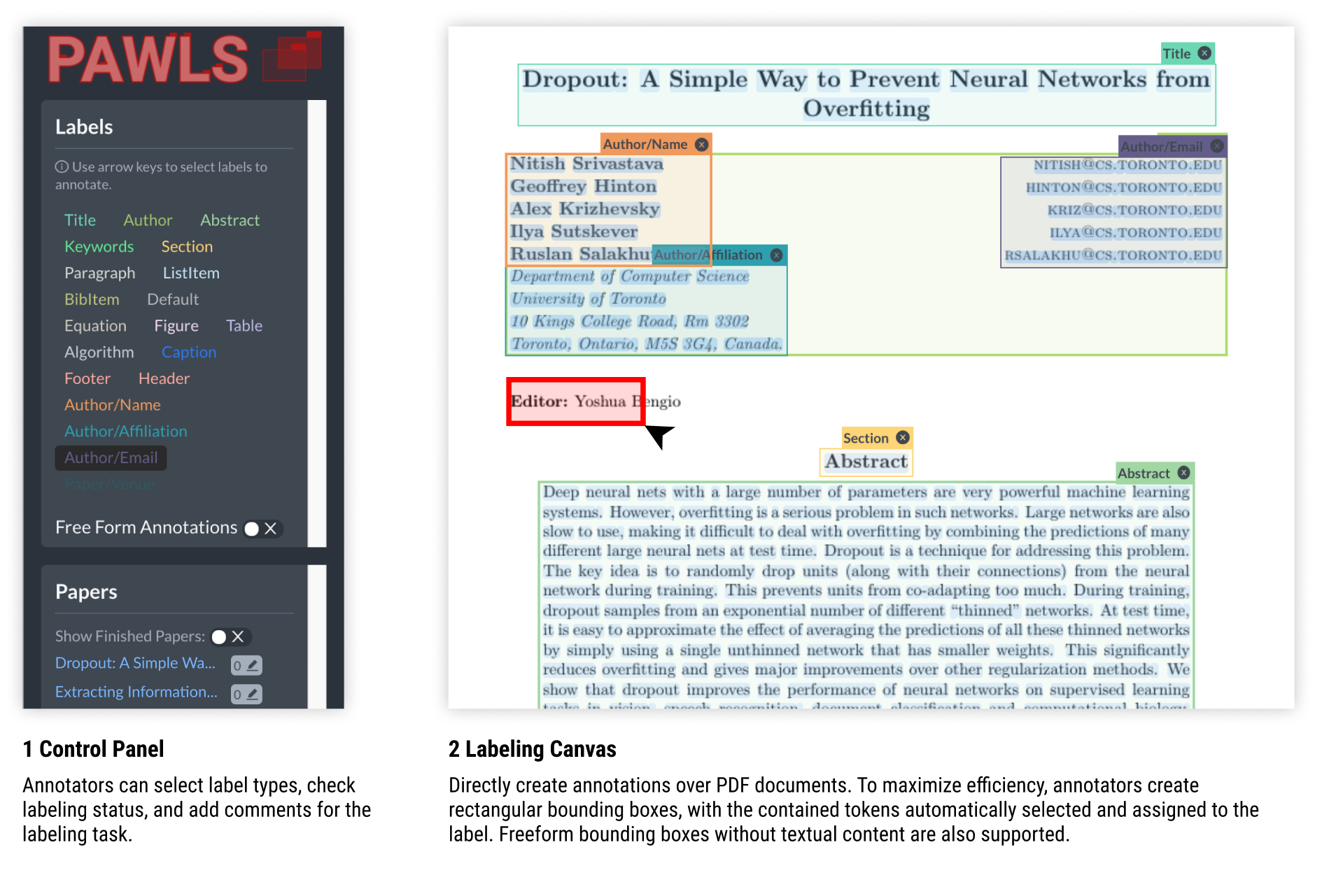}
    \caption{An overview of the \pawls annotation interface.}
    \label{fig:overview}
\end{figure*}

\section{Design Choices}

As shown in Figure~\ref{fig:overview}, the primary operation that \pawls supports is drawing a bounding box over a PDF document with the mouse, and assigning that region of the document a textual label. \pawls supports drawing both freeform boxes anywhere on the PDF, as well as boxes which are associated with tokens extracted from the PDF itself.

This section describes some of the user interface design choices in \pawls.

\subsection{PDF Native Annotation}

The primary tenet of \pawls is the idea that annotators are accustomed to reading and interacting with PDF documents themselves, and as such, \pawls should render the actual PDF as the medium for annotation. In order to achieve this, annotations themselves must be relative to a rendered PDF's scale in the browser. Annotations are automatically re-sized to fit the rendered PDF canvas, but stored relative to the absolute dimensions of the original PDF document.

\subsection{Annotator Ease of Use}

\pawls contains several features which are designed to speed up annotation by users, as well as minimizing frustrating or difficult interaction experiences. Bounding box borders in \pawls change depending on the size and density of the annotated span, making it easier to read dense annotations. Annotators can hide bounding box labels using the CTRL key for cases where labels are obscuring the document flow. Users can undo annotations with familiar key combinations (CMD-z) and delete annotations directly from the sidebar. These features were derived from a tight feedback loop with annotation experts during development of the tool. 

\begin{figure*}[ht]
    \centering
    \includegraphics[width=\textwidth]{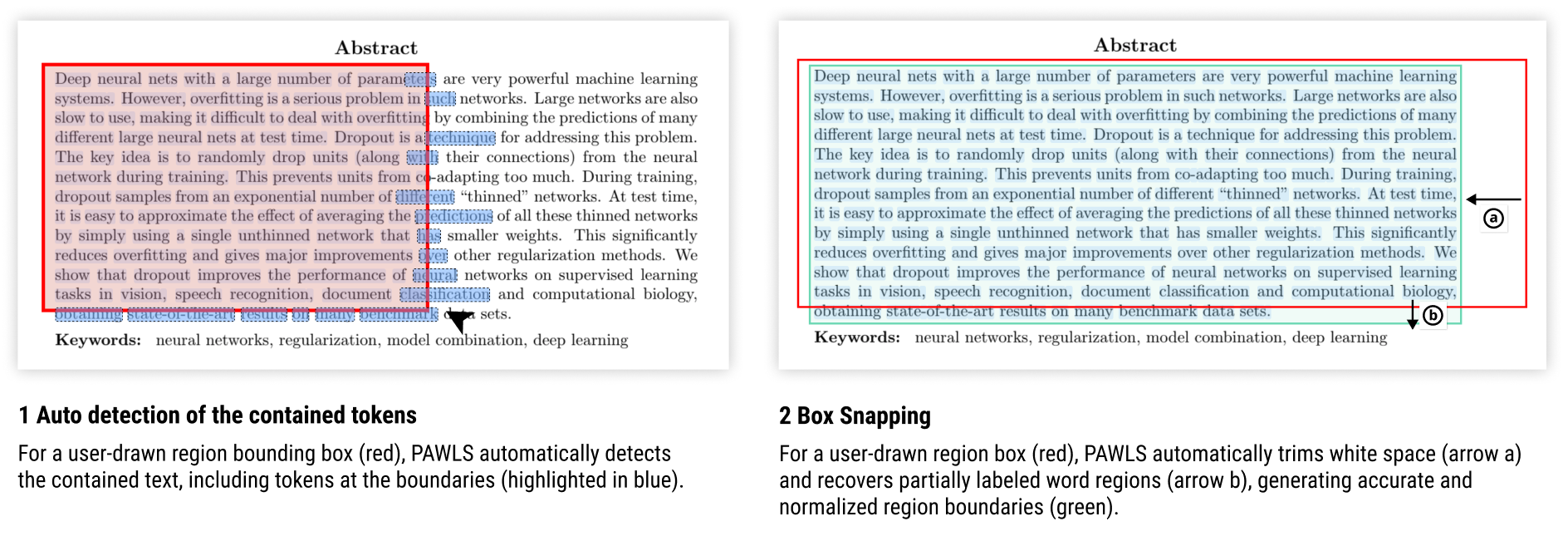}
    \caption{An example of visual token selection. When a user begins highlighting a bounding box, \pawls uses underlying token level boundary information extracted from the PDF to 1) highlight selected textual spans as they are dragged over and 2) normalize the bounding box of a selection to be a fixed padded distance from the maximally large token boundary.}
    \label{fig:bounding_box}
\end{figure*}

\subsection{Token Parsing}

\pawls pre-processes PDFs before they are rendered in the UI to extract the bounding boxes of every token present in the document. This allows a variety of interactive labelling features described below. Users can choose between different pre-processors based on their needs, such as GROBID \footnote{https://github.com/kermitt2/grobid} and PdfPlumber \footnote{https://github.com/jsvine/pdfplumber} for programmatically generated PDFs, or Tesseract \footnote{https://github.com/tesseract-ocr/tesseract} for Optical Character Recognition (OCR) in PDFs which have been scanned, or are otherwise low quality. Future extensions to \pawls will include higher level PDF structure which is general enough to be useful across a range of domains, such as document titles, paragraphs and section headings to further extend the possible annotation modes, such as clicking on paragraphs or sections.

\subsection{Visual Token Selection and Box Snapping}

\pawls pre-processes PDFs before they are served in the annotation interface, giving access to token level bounding box information. When users draw new bounding boxes, token spans are highlighted to indicate their inclusion in the annotation. After the user has completed the selection, the bounding box ``snaps'' to a normalized boundary containing the underlying PDF tokens. Figure \ref{fig:bounding_box} demonstrates this interaction. In particular, this allows bounding boxes to be normalized relative to their containing token positions (having a fixed border), making annotations more consistent and uniform with no additional annotator effort. This feature allows annotators to focus on the content of their annotations, rather than ensuring a consistent visual markup, easing the annotation flow and increasing the consistency of the collected annotations.

\subsection{N-ary Relational Annotations}

\begin{figure}[ht]
    \centering
    \includegraphics[width=0.5\textwidth]{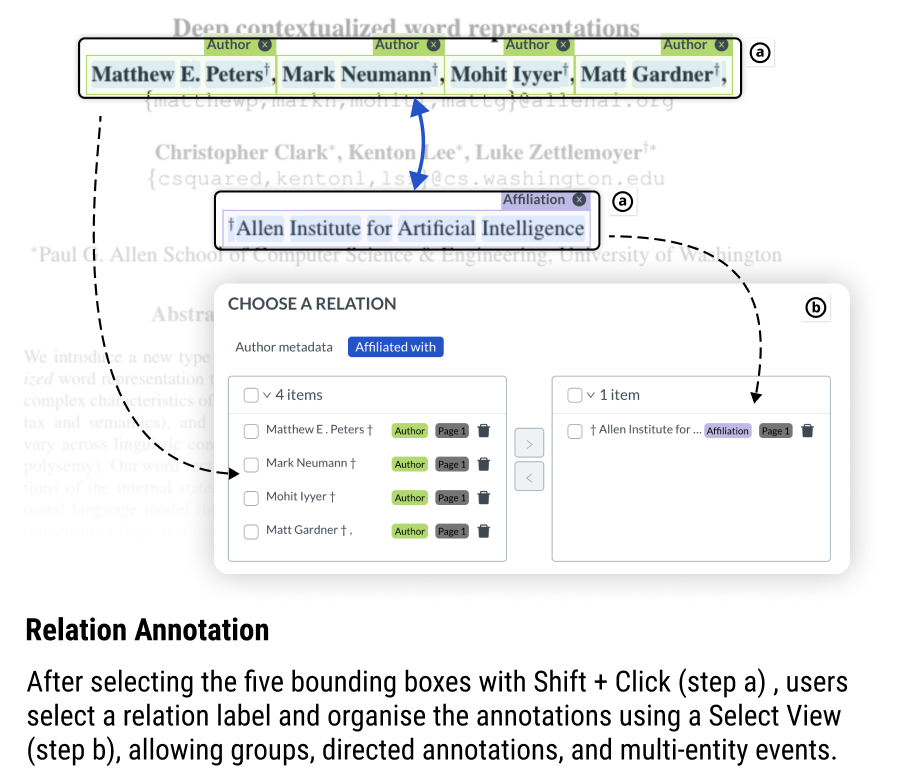}
    \caption{The relation annotation modal.}
    \label{fig:relations}
\end{figure}

\pawls supports N-ary relational annotations as well as those based on bounding boxes. Relational annotations are supported for both textual and free-form annotations, allowing the collection of event structures which include non-textual PDF regions, such as figure/table references, or sub-image coordination. For example, this feature would allow annotators to link figure captions to particular figure regions, or relate a discussion of a particular table column in the text to the exact visual region of the column/table itself. Figure \ref{fig:relations} demonstrates this interaction mode for two annotations.

\subsection{Command Line Interface}

\pawls includes a command line interface for administrating annotation projects. It includes functionality for assigning labeling tasks to annotators, monitoring the annotation progress and quality (measuring inter annotator agreement), and exporting annotations in a variety of formats. Additionally, it supports pre-populating annotations from model predictions, detailed in Section~\ref{sec:pre_population}. 

Annotations in \pawls can be exported to different formats to support different downstream tasks. The hierarchical structure of user-drawn blocks and PDF tokens is stored in JSON format, linking blocks with their corresponding tokens. For vision-centered tasks (e.g., document layout detection), \pawls supports converting to the widely-used COCO format, including generating jpeg captures of pdf pages for training vision models. For text-centric tasks, \pawls can generate a table for tokens and labels obtained from the annotated bounding boxes.

\subsection{Annotation Pre-population}
\label{sec:pre_population}

The \pawls command line interface supports pre-population of annotations given a set of bounding boxes predictions for each page. This enables model-in-the-loop type functionality, with annotators correcting model predictions directly on the PDF. Future extensions to \pawls will include active learning based annotation suggestions as annotators work, from models running as a service.

\section{Implementation}

\pawls is implemented as a Python-based web server which serves PDFs, annotations and other metadata stored on disk in the JSON format. The user interface is a Single Page Application implemented using Typescript and relies heavily on the React web framework. PDFs are rendered using PDF.js. \pawls is designed to be used in a browser, with no setup work required on the behalf of annotators apart from navigating to a web page. This makes annotation projects more flexible as they can be distributed across a variety of crowd-sourcing platforms, used in house, or run on local machines.

\pawls development and deployment are both managed using the containerization tools Docker and Docker Compose, and multiple \pawls instances are running on a Google Cloud Platform Kubernetes cluster. Authentication in production environments is managed via Google Account logins, but \pawls can be run locally by individual users with no authentication. 

\section{Case Study}

\pawls enables the collection of mixed-mode annotations on PDFs. \pawls is currently in use for a PDF Layout Parsing project for academic papers, for which we have collected an initial set of gold standard annotations. This dataset consists of 80 PDF pages with 2558 densely annotated bounding boxes of 20 categories from 3 annotators.

Table~\ref{table:acc} reports pairwise Inter-Annotator agreement scores, split out into textual and non-textual labels. For textual labels like titles and paragraphs, the agreement is measured via token accuracy: for each word labeled, we compare the label of the belonging block across different annotators. Non-textual labels are used for regions like figures and tables, and they are usually labeled using free-form boxes. Average Precision (AP) score~\cite{lin2014microsoft}, commonly used in Object Detection tasks (e.g., COCO) in computer vision, is adopted to measure the consistency of these boxes labeled by different annotators. As AP calculates the block categories agreement at different overlapping levels, the scoring is not commutative.

\begin{table}[h]
    \resizebox{1.\linewidth}{!}{
    \begin{threeparttable}
        \begin{tabular}{r|ccc}
        \toprule
                    & \textbf{Annotator 1}   & \textbf{Annotator 2}
                    & \textbf{Annotator 3}   \\
        \midrule
        \textbf{Annotator 1} & N/A & 94.43 / 86.58 & 93.28 / 83.97 \\
        \textbf{Annotator 2} & 94.43 / 86.49 & N/A & 88.69 / 84.20 \\
        \textbf{Annotator 3} & 93.28 / 84.67 & 88.69 / 84.79 & N/A               \\
        \bottomrule
        \end{tabular}
    \end{threeparttable}
    }
\caption{The Inter-Annotator Agreement scores for the labeling task. We show the textual / non-textual annotation agreement scores in each cell. The $(i,j)$-th element in this table is calculated by treating $i$'s annotation as the ``ground truth'' and $j$'s as the ``prediction''.}
\label{table:acc}
\end{table}

\section{Related Work}

Many commercial PDF annotation tools exist, such as IBM Watson's smart document understanding feature and TagTog's Beta PDF Annotation tool \footnote{\url{https://www.tagtog.net/\#pdf-annotation}}. \pawls will be open source and freely available. Knowledge management systems such as Prot{\'e}g{\'e} \cite{Musen2015ThePP} support PDFs, but more suited to management of large, evolving corpora and knowledge graph construction than the creation of static datasets.

LabelStudio \footnote{\url{https://labelstud.io/}} supports image annotation as well as plaintext/html-based annotation, meaning PDF pages can be uploaded and annotated within their user interface. However, bounding boxes are hand drawn, and the context of the entire PDF is not visible as the pdf pages are viewed as individual images. PDFAnno \cite{Shindo2018PDFAnnoAW} is the closest tool conceptually to \pawls, supporting multiple annotation modes and pdf-based rendering. Unfortunately PDFAnno is no longer maintained and \pawls provides additional functionality, such as pre-annotation.

Several PDF based datasets exist for document parsing, such as DocBank \cite{Li2020DocBankAB}, PubLeNet \cite{Zhong2019PubLayNetLD} and TableBank \cite{Li2019TableBankAB}. However, both DocBank and PubLeNet are constructed using weak supervision from Latex parses or Pubmed XML information. TableBank consists of 417k tables extracted from Microsoft Word documents and computer generated PDFs. This approach is feasible for common elements of document structure such as tables, but is not possible for custom annotation labels or detailed figure/table decomposition.

The \pawls interface is similar to tools which augment PDFs for reading or note taking purposes. Along with commercial tools such as Adobe Reader, SideNoter \cite{abekawa-aizawa-2016-sidenoter} augments PDFs with rich note taking and linguistic annotation overlays, directly on the PDF canvas. ScholarPhi \cite{Head2020AugmentingSP} augments the PDF reading experience with equation overlays and definition modals for symbols.

As a PDF specific annotation tool, \pawls adds to the wider landscape of annotation tools which fulfil a particular niche. SLATE \cite{kummerfeld-2019-slate} provides a command line annotation tool for expert annotators; \cite{mayhew-roth-2018-talen} provides an annotation interface specifically designed for cross-lingual annotation in which the annotators do not speak the target language.
 
Textual annotation tools such as BRAT \cite{stenetorp-etal-2012-brat}, Pubtator \cite{10.1093/nar/gkt441, pubtator} or Knowtator \cite{Ogren2006KnowtatorAP} are recommended for annotations which do not require full PDF context, or for which extension to multi-modal data formats is not possible or likely. We view \pawls as a complimentary tool to the suite of text based annotation tools, which support more advanced types of annotation and configuration, but deal with annotation on extracted text removed from it's originally published format.

In particular, we envisage scholarly document annotation as a key use case for \pawls, as PDF is a widely used format in the context of scientific publication. Several recently published datasets leave document structure parsing or multi-modal annotation to future work. For example, the SciREX dataset \cite{jain-etal-2020-scirex} use the text-only LaTeX source of ArXiv papers for dataset construction, leaving Table and Figure extraction to future work. Multiple iterations of the Evidence Inference dataset \cite{lehman2019inferring, deyoung2020evidence} use textual descriptions of interventions in clinical trial reports; answering inferential questions using figures, tables and graphs may be a more natural format for some queries.

\section{Conclusion}

In this paper, we have introduced a new annotation tool, \pawls, designed specifically with PDFs in mind. \pawls facilitates the creation of multi-modal datasets, due to its support for mixed mode annotation of both text and image sub-regions on PDFs. Additionally, we described several user interface design choices which improve the resulting annotation quality, and conducted a small initial annotation effort, reporting high annotator agreement. \pawls is released as an open source project under the Apache 2.0 license.

\bibliography{naacl2021}

\begin{thebibliography}{21}
\expandafter\ifx\csname natexlab\endcsname\relax\def\natexlab#1{#1}\fi

\bibitem[{Abekawa and Aizawa(2016)}]{abekawa-aizawa-2016-sidenoter}
Takeshi Abekawa and Akiko Aizawa. 2016.
\newblock \href {https://www.aclweb.org/anthology/C16-2029} {{S}ide{N}oter:
  Scholarly paper browsing system based on {PDF} restructuring and text
  annotation}.
\newblock In \emph{Proceedings of {COLING} 2016, the 26th International
  Conference on Computational Linguistics: System Demonstrations}, pages
  136--140, Osaka, Japan. The COLING 2016 Organizing Committee.

\bibitem[{Devlin et~al.(2019)Devlin, Chang, Lee, and
  Toutanova}]{devlin-etal-2019-bert}
Jacob Devlin, Ming-Wei Chang, Kenton Lee, and Kristina Toutanova. 2019.
\newblock \href {https://doi.org/10.18653/v1/N19-1423} {{BERT}: Pre-training of
  deep bidirectional transformers for language understanding}.
\newblock In \emph{Proceedings of the 2019 Conference of the North {A}merican
  Chapter of the Association for Computational Linguistics: Human Language
  Technologies, Volume 1 (Long and Short Papers)}, pages 4171--4186,
  Minneapolis, Minnesota. Association for Computational Linguistics.

\bibitem[{DeYoung et~al.(2020)DeYoung, Lehman, Nye, Marshall, and
  Wallace}]{deyoung2020evidence}
Jay DeYoung, Eric Lehman, Ben Nye, Iain~J. Marshall, and Byron~C. Wallace.
  2020.
\newblock \href {http://arxiv.org/abs/2005.04177} {Evidence inference 2.0: More
  data, better models}.

\bibitem[{Head et~al.(2020)Head, Lo, Kang, Fok, Skjonsberg, Weld, and
  Hearst}]{Head2020AugmentingSP}
Andrew Head, Kyle Lo, Dongyeop Kang, Raymond Fok, Sam Skjonsberg, Daniel~S.
  Weld, and Marti~A. Hearst. 2020.
\newblock Augmenting scientific papers with just-in-time, position-sensitive
  definitions of terms and symbols.
\newblock \emph{ArXiv}, abs/2009.14237.

\bibitem[{Jain et~al.(2020)Jain, van Zuylen, Hajishirzi, and
  Beltagy}]{jain-etal-2020-scirex}
Sarthak Jain, Madeleine van Zuylen, Hannaneh Hajishirzi, and Iz~Beltagy. 2020.
\newblock \href {https://doi.org/10.18653/v1/2020.acl-main.670} {{S}ci{REX}:
  {A} challenge dataset for document-level information extraction}.
\newblock In \emph{Proceedings of the 58th Annual Meeting of the Association
  for Computational Linguistics}, pages 7506--7516, Online. Association for
  Computational Linguistics.

\bibitem[{Joseph et~al.(2017)Joseph, Friedland, Hobbs, Lazer, and
  Tsur}]{joseph-etal-2017-constance}
Kenneth Joseph, Lisa Friedland, William Hobbs, David Lazer, and Oren Tsur.
  2017.
\newblock \href {https://doi.org/10.18653/v1/D17-1116} {{C}on{S}tance: Modeling
  annotation contexts to improve stance classification}.
\newblock In \emph{Proceedings of the 2017 Conference on Empirical Methods in
  Natural Language Processing}, pages 1115--1124, Copenhagen, Denmark.
  Association for Computational Linguistics.

\bibitem[{Kummerfeld(2019)}]{kummerfeld-2019-slate}
Jonathan~K. Kummerfeld. 2019.
\newblock \href {https://doi.org/10.18653/v1/P19-3002} {{SLATE}: A
  super-lightweight annotation tool for experts}.
\newblock In \emph{Proceedings of the 57th Annual Meeting of the Association
  for Computational Linguistics: System Demonstrations}, pages 7--12, Florence,
  Italy. Association for Computational Linguistics.

\bibitem[{L{\"a}ubli et~al.(2018)L{\"a}ubli, Sennrich, and
  Volk}]{laubli-etal-2018-machine}
Samuel L{\"a}ubli, Rico Sennrich, and Martin Volk. 2018.
\newblock \href {https://doi.org/10.18653/v1/D18-1512} {Has machine translation
  achieved human parity? a case for document-level evaluation}.
\newblock In \emph{Proceedings of the 2018 Conference on Empirical Methods in
  Natural Language Processing}, pages 4791--4796, Brussels, Belgium.
  Association for Computational Linguistics.

\bibitem[{Lehman et~al.(2019)Lehman, DeYoung, Barzilay, and
  Wallace}]{lehman2019inferring}
Eric Lehman, Jay DeYoung, Regina Barzilay, and Byron~C Wallace. 2019.
\newblock Inferring which medical treatments work from reports of clinical
  trials.
\newblock In \emph{Proceedings of the North American Chapter of the Association
  for Computational Linguistics (NAACL)}, pages 3705--3717.

\bibitem[{Li et~al.(2020{\natexlab{a}})Li, Cui, Huang, Wei, Zhou, and
  Li}]{Li2019TableBankAB}
Minghao Li, Lei Cui, Shaohan Huang, Furu Wei, Ming Zhou, and Zhoujun Li.
  2020{\natexlab{a}}.
\newblock \href {https://www.aclweb.org/anthology/2020.lrec-1.236}
  {{T}able{B}ank: Table benchmark for image-based table detection and
  recognition}.
\newblock In \emph{Proceedings of the 12th Language Resources and Evaluation
  Conference}, pages 1918--1925, Marseille, France. European Language Resources
  Association.

\bibitem[{Li et~al.(2020{\natexlab{b}})Li, Xu, Cui, Huang, Wei, Li, and
  Zhou}]{Li2020DocBankAB}
Minghao Li, Yiheng Xu, Lei Cui, Shaohan Huang, Furu Wei, Zhoujun Li, and
  M.~Zhou. 2020{\natexlab{b}}.
\newblock Docbank: A benchmark dataset for document layout analysis.
\newblock \emph{ArXiv}, abs/2006.01038.

\bibitem[{Lin et~al.(2014)Lin, Maire, Belongie, Hays, Perona, Ramanan,
  Doll{\'a}r, and Zitnick}]{lin2014microsoft}
Tsung-Yi Lin, Michael Maire, Serge Belongie, James Hays, Pietro Perona, Deva
  Ramanan, Piotr Doll{\'a}r, and C~Lawrence Zitnick. 2014.
\newblock Microsoft coco: Common objects in context.
\newblock In \emph{European conference on computer vision}, pages 740--755.
  Springer.

\bibitem[{Mayhew and Roth(2018)}]{mayhew-roth-2018-talen}
Stephen Mayhew and Dan Roth. 2018.
\newblock \href {https://doi.org/10.18653/v1/P18-4014} {{TALEN}: Tool for
  annotation of low-resource {EN}tities}.
\newblock In \emph{Proceedings of {ACL} 2018, System Demonstrations}, pages
  80--86, Melbourne, Australia. Association for Computational Linguistics.

\bibitem[{Musen(2015)}]{Musen2015ThePP}
M.~Musen. 2015.
\newblock The prot{\'e}g{\'e} project: a look back and a look forward.
\newblock \emph{AI matters}, 1 4:4--12.

\bibitem[{Ogren(2006)}]{Ogren2006KnowtatorAP}
Philip~V. Ogren. 2006.
\newblock Knowtator: A prot{\'e}g{\'e} plug-in for annotated corpus
  construction.
\newblock In \emph{HLT-NAACL}.

\bibitem[{Peters et~al.(2018)Peters, Neumann, Iyyer, Gardner, Clark, Lee, and
  Zettlemoyer}]{peters-etal-2018-deep}
Matthew Peters, Mark Neumann, Mohit Iyyer, Matt Gardner, Christopher Clark,
  Kenton Lee, and Luke Zettlemoyer. 2018.
\newblock \href {https://doi.org/10.18653/v1/N18-1202} {Deep contextualized
  word representations}.
\newblock In \emph{Proceedings of the 2018 Conference of the North {A}merican
  Chapter of the Association for Computational Linguistics: Human Language
  Technologies, Volume 1 (Long Papers)}, pages 2227--2237, New Orleans,
  Louisiana. Association for Computational Linguistics.

\bibitem[{Shindo et~al.(2018)Shindo, Munesada, and
  Matsumoto}]{Shindo2018PDFAnnoAW}
Hiroyuki Shindo, Yohei Munesada, and Y.~Matsumoto. 2018.
\newblock Pdfanno: a web-based linguistic annotation tool for pdf documents.
\newblock In \emph{LREC}.

\bibitem[{Stenetorp et~al.(2012)Stenetorp, Pyysalo, Topi{\'c}, Ohta, Ananiadou,
  and Tsujii}]{stenetorp-etal-2012-brat}
Pontus Stenetorp, Sampo Pyysalo, Goran Topi{\'c}, Tomoko Ohta, Sophia
  Ananiadou, and Jun{'}ichi Tsujii. 2012.
\newblock \href {https://www.aclweb.org/anthology/E12-2021} {brat: a web-based
  tool for {NLP}-assisted text annotation}.
\newblock In \emph{Proceedings of the Demonstrations at the 13th Conference of
  the {E}uropean Chapter of the Association for Computational Linguistics},
  pages 102--107, Avignon, France. Association for Computational Linguistics.

\bibitem[{Wei et~al.(2012)Wei, Kao, and Lu}]{pubtator}
Chih-Hsuan Wei, Hung-Yu Kao, and Zhiyong Lu. 2012.
\newblock Pubtator: A pubmed-like interactive curation system for document
  triage and literature curation.
\newblock \emph{BioCreative 2012 workshop}, 05.

\bibitem[{Wei et~al.(2013)Wei, Kao, and Lu}]{10.1093/nar/gkt441}
Chih-Hsuan Wei, Hung-Yu Kao, and Zhiyong Lu. 2013.
\newblock \href {https://doi.org/10.1093/nar/gkt441} {Pubtator: a web-based
  text mining tool for assisting biocuration}.
\newblock \emph{Nucleic Acids Research}, 41.

\bibitem[{Zhong et~al.(2019)Zhong, Tang, and
  Jimeno-Yepes}]{Zhong2019PubLayNetLD}
Xu~Zhong, J.~Tang, and Antonio Jimeno-Yepes. 2019.
\newblock Publaynet: Largest dataset ever for document layout analysis.
\newblock \emph{2019 International Conference on Document Analysis and
  Recognition (ICDAR)}, pages 1015--1022.

\end{thebibliography}
\bibliographystyle{acl_natbib}

\appendix

\section{Accessing the Demo} \label{demo_access}

Production deployments of \pawls use Google Authentication to allow users to log in. The demo server, accessible at \url{https://pawls.apps.allenai.org/}, is configured to allow access to all non-corp gmail accounts, e.g any email address ending in ``@gmail.com". No annotations will be collected from this server, as it is read-only. Please use a personal email address, or create a one-off account if you do not use gmail. If you cannot log in, please try again using an incognito window which will ensure gmail cookies are not set. A demo video for \pawls usage is available at \url{https://youtu.be/TB4kzh2H9og}.

\end{document}